# Deep Representation Learning for Electronic Design Automation

Pratik Shrestha
Drexel University
Philadelphia, Pennsylvania, USA
ps937@drexel.com

Saran Phatharodom
Drexel University
Philadelphia, Pennsylvania, USA
sp694@drexel.com

Alec Aversa
Drexel University
Philadelphia, Pennsylvania, USA
aja367@drexel.com

David Blankenship
Drexel University
Philadelphia, Pennsylvania, USA
dwb65@drexel.com

Jeff Wu
Drexel University
Philadelphia, Pennsylvania, USA
jw3723@drexel.com

Ioannis Savidis
Drexel University
Philadelphia, Pennsylvania, USA
is338@drexel.com

## ABSTRACT

Representation learning has become an effective technique utilized by electronic design automation (EDA) algorithms, which leverage the natural representation of workflow elements as images, grids, and graphs. By addressing challenges related to the increasing complexity of circuits and stringent power, performance, and area (PPA) requirements, representation learning facilitates the automatic extraction of meaningful features from complex data formats, including images, grids, and graphs. This paper examines the application of representation learning in EDA, covering foundational concepts and analyzing prior work and case studies on tasks that include timing prediction, routability analysis, and automated placement. Key techniques, including image-based methods, graph-based approaches, and hybrid multimodal solutions, are presented to illustrate the improvements provided in routing, timing, and parasitic prediction. The provided advancements demonstrate the potential of representation learning to enhance efficiency, accuracy, and scalability in current integrated circuit design flows.

## 1 INTRODUCTION

The automated design flow of digital circuits, which is shown in Fig. 1, transforms hardware design language (HDL) into manufacturable integrated circuits while meeting power, performance, and area (PPA) specifications through stages that include logical synthesis, floorplanning, placement, clock network synthesis, and routing. The stages rely on technology-specific constraints and generate circuit descriptors that include netlists, interconnect parasitics, and reports to validate the design against target PPA objectives. However, increasing circuit complexity, technology scaling, and stringent PPA requirements challenge traditional electronic design automation (EDA) tools, often resulting in suboptimal designs and failed timing closure that require manual adjustments to the circuit design and fabrication delays [1]. Machine learning (ML) is, therefore, currently being pursued to enhance efficiency across the design flow through predictive modeling, adaptive optimization, and automation [2].

Artifacts, metrics, and data produced during the automated design flow can be represented as images or graphs, which has resulted in an increased use of image- and graph-based learning techniques within EDA workflows. Representation learning [3] represents a promising direction for advancing EDA methodologies. By extracting meaningful features from raw data, deep representation learning effectively captures spatial, hierarchical, and relational patterns inherent in graph-based netlists and image-based layouts.

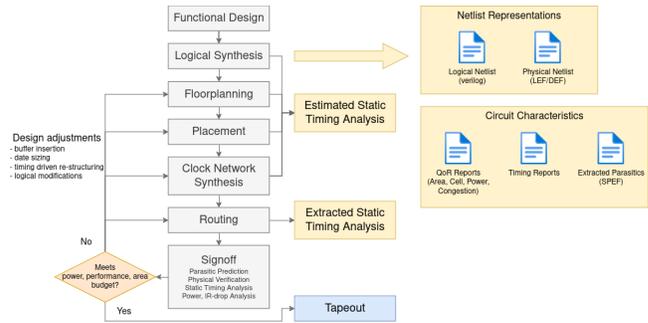

Figure 1: Block representation of design automation flow.

This paper provides an overview of representation learning algorithms and the application of such algorithms in EDA. Foundational concepts of representation learning and key methodologies relevant to EDA are introduced in Section 2. Prior work that utilizes representation learning within the EDA flow and three case studies exploring timing analysis, routability prediction, and automated placement are described in Section 3. Potential future directions identifying opportunities for advancements in representation learning to address unresolved challenges in EDA are presented in Section 4. Finally, some concluding remarks are provided in Section 5.

## 2 BACKGROUND ON REPRESENTATION LEARNING

Representation learning, a subset of machine learning, encompasses algorithms designed to uncover underlying patterns or representations within complex data. High-dimensional raw data are transformed into low-dimensional vector representations, or "embeddings," that preserve essential features, disentangle key characteristics, and minimize irrelevant noise. The embeddings simplify the identification of complex patterns and relationships, which enables the utilization of embeddings in downstream tasks that include classification, clustering, and regression.

### 2.1 Modality Embedding

In representation learning, the term *modality* refers to the type of data being processed. Common modalities include text, images, audio, video, and graphs. Each modality requires tailored approaches for effective representation learning, which ensures that the unique structure and features of the data are captured. This section explores the methods and network architectures specific to key modalities.

*2.1.1 Images.* Image and grid-based data, commonly represented as arrays of pixels or numerical grids, are data formats naturally suited for convolutional neural networks (CNNs) [4] and excel at extracting spatial patterns and hierarchical representations.

CNNs operate by applying convolutional filters that slide across an input grid or image, producing feature maps that capture localized patterns including edges, textures, and shapes. The operation of a convolutional layer is mathematically defined as

$$\vec{y}_{i,j} = \sum_{m=0}^{M-1} \sum_{n=0}^{N-1} \vec{x}_{i+m,j+n} \cdot \vec{w}_{m,n} + b, \quad (1)$$

where $\vec{x}$ represents the input data (e.g., pixel values or grid elements), $\vec{w}$ is the filter kernel of dimensions $M \times N$, $b$ is the bias term, and $\vec{y}$ is the resulting feature map. $i$ and $j$ denote the row and column indices of the output feature map ($\vec{y}$). Through successive neural network layers, CNNs extract increasingly abstract features, enabling the network to effectively capture meaningful representations of the input data.

Residual networks (ResNet) [5] are a widely used family of architectures that address challenges in training deep neural networks, including the vanishing gradient problem, by introducing residual connections, a connection that allows input to bypass one or more layers, which ensures efficient gradient flow. Networks such as ResNet-18, ResNet-50, and ResNet-101 are architectures utilized for the processing of image- and grid-based data. Pre-trained models on datasets such as ImageNet [6] facilitate transfer learning, reduce training time, and enhance performance in data-rich domains.

*2.1.2 Graphs.* Graphs provide a natural representation of structured data, capturing complex relationships and interactions between entities as nodes and edges. Graph neural networks (GNNs) are essential for embedding and analyzing such data to extract meaningful representations. Among various types of GNNs, Graph Convolutional Networks (GCNs) [7] operate on graph data by aggregating features from the neighbors of a given node. The embedding of each node is updated by combining information from the local neighborhood of the given node. The embedding operation between neural network layers is given by

$$\mathbf{H}^{(l+1)} = \sigma(\tilde{\mathbf{D}}^{-\frac{1}{2}} \tilde{\mathbf{A}} \tilde{\mathbf{D}}^{-\frac{1}{2}} \mathbf{H}^{(l)} \mathbf{W}^{(l)}), \quad (2)$$

where $\mathbf{H}^{(l)}$ is the feature matrix at layer $l$ of the neural network, $\tilde{\mathbf{A}} = \mathbf{A} + \mathbf{I}$ is the adjacency matrix $\mathbf{A}$ with self-loops provided by identity matrix $\mathbf{I}$, $\tilde{\mathbf{D}}$ is the degree matrix of $\tilde{\mathbf{A}}$, $\mathbf{W}^{(l)}$ is the learnable weight matrix, and $\sigma$ is an activation function.

Graph Attention Networks (GATs) [8] extend GCNs by incorporating attention mechanisms to dynamically weigh the importance of neighboring nodes during aggregation, similar to how transformers [9] use attention to capture relationships across input data. GATs are, therefore, able to target the most relevant neighbors of each node. The updated embedding for a node is computed as

$$\vec{h}_i^{(l+1)} = \sigma\left(\sum_{j \in \mathcal{N}(i)} \alpha_{ij} \mathbf{W}^{(l)} \vec{h}_j^{(l)}\right), \quad (3)$$

where $\alpha_{ij}$ represents the attention coefficient between nodes $i$ and $j$, $\mathcal{N}(i)$ represents the set of neighboring nodes of node $i$ in the graph, $\mathbf{W}^{(l)}$ is the learnable weight matrix, and $\vec{h}_j^{(l)}$ is the feature vector of neighbor $j$. The attention coefficient is defined as

$$\alpha_{ij} = \frac{\exp\left(\text{LeakyReLU}\left(\vec{a}^\top [\mathbf{W}^{(l)} \vec{h}_i^{(l)} \| \mathbf{W}^{(l)} \vec{h}_j^{(l)}]\right)\right)}{\sum_{k \in \mathcal{N}(i)} \exp\left(\text{LeakyReLU}\left(\vec{a}^\top [\mathbf{W}^{(l)} \vec{h}_i^{(l)} \| \mathbf{W}^{(l)} \vec{h}_k^{(l)}]\right)\right)}, \quad (4)$$

where $\vec{a}$ is a learnable vector and $\|$ denotes concatenation [8].

GraphSAGE [10] samples and aggregates information from a fixed subset of neighbors $\mathcal{S}(i) \subseteq \mathcal{N}(i)$, explicitly selecting a subset of neighbors for each node $i$ at every layer, making the embedding layer scalable for large and dynamic graphs. The aggregation is defined as:

$$\vec{h}_i^{(l+1)} = \sigma\left(\mathbf{W}^{(l)} \cdot \text{AGG}\left(\{\vec{h}_j^{(l)} \mid j \in \mathcal{S}(i)\}\right)\right), \quad (5)$$

where AGG is the aggregation function (e.g., mean, max, or sum), $\mathbf{W}^{(l)}$ is the learnable weight matrix, and $\vec{h}_j^{(l)}$ represents the feature vector of neighbor $j$.

GCNs, GATs, and GraphSAGE differ in the approach used to aggregate information from graph structures, each offering unique strengths and trade-offs. GCNs aggregate features uniformly, ensuring computational efficiency, but assume equal importance to all neighbors of a node, which limits the ability to capture complex relationships. GATs, in contrast, introduce attention mechanisms to dynamically weigh the contributions of each neighbor, allowing the model to better utilize the most relevant nodes. Attention improves performance, but adds computational overhead. GraphSAGE explicitly samples a fixed subset of neighbors for each node, trading off exhaustively considering neighbors for scalability in large and dynamic graphs. Together, the three methods highlight the trade-offs between computational complexity, scalability, and the ability to capture nuanced graph relationships.

## 2.2 Autoencoders

Autoencoders[11] are a type of artificial neural network designed to learn efficient data representations in an unsupervised manner. An example autoencoder is shown in Fig. 2a, which consists of two primary components: an encoder and a decoder. The encoder compresses input into a reduced latent space, retaining key features, while the decoder decompresses the latent space to match the original data. Both the encoder and decoder are composed of feedforward layers with activation functions that transform data into and out of the latent space. The latent representation within the bottleneck of the autoencoder acts as a compressed form of the data, capturing the most essential patterns and features of the data, which makes autoencoders useful for downstream prediction tasks.

U-Nets [12], specialized autoencoders structured on an encoder-decoder architecture, contain a contracting path for capturing context from the input and an expansive path for constructing an output from the latent space. An example U-net is shown in Fig. 2b. The encoding path is comprised of down-sampling convolution layers, and the decoding path is comprised of up-sampling de-convolution layers. In addition, U-Nets include skip connections that link corresponding layers in the encoder and decoder paths, enabling the direct transfer of fine-grained details to improve localization and the accuracy of the reconstructed data.

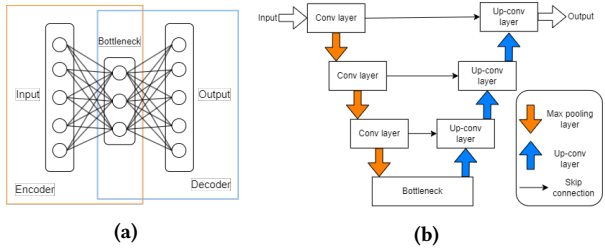

**Figure 2: (a) Representation of an autoencoder with encoder, bottleneck, and decoder for data compression and reconstruction and (b) a U-Net with encoder-decoder paths and skip connections for improved reconstruction accuracy.**

## 3 USE CASES OF REPRESENTATION LEARNING IN EDA

The practical applications of deep representation learning techniques in EDA are highlighted in Section 3. Prior work that utilizes ML for routability prediction, timing analysis, and automated placement, which are explored in detail in Sections 3.1, 3.2, and 3.3, respectively, are summarized in Table 1. For each general use case, the table is categorized by the problem type (e.g., classification, regression, generative), data modality (e.g., images, graphs), representation techniques, and the neural network architectures employed. The body of work that falls under each of the described cases highlight targeted applications of representation learning techniques that address specific challenges in EDA workflows.

### 3.1 Case I: Routability Prediction

Routability is a measure of how effectively signal paths are routed between modules and devices within a circuit while adhering to design constraints. Congested areas within a circuit are susceptible to violating design rules and degrading performance due to high-density and high-activity hotspots. Predicting routing congestion is essential in order to identify high-risk zones within an integrated circuit (IC). Utilizing predicted routing congestion to optimize the placement of cells and modules by identifying overflow areas and improving routing efficiency ensures the circuit is both functional and manufacturable. Image-based machine learning is generally used for such predictions, RouteNet [13] being one of the earliest examples. RouteNet introduces a dual-model approach, utilizing a CNN (ResNet) to predict design rule violations (DRVs) and a U-net to detect areas of a circuit considered design rule check (DRC) hotspots. Using image-like representations of layout data (e.g., pin density, macro regions, RUDY [33] scores, etc.), RouteNet matches the accuracy of traditional global routing algorithms while being significantly faster and providing improved accuracy of detected DRC hotspots. J-nets are introduced in [14], which is an extension of the U-net, for predicting post-routing DRVs. J-nets utilize post-placement layout images as inputs, combining high-resolution pin configuration images with low-resolution tile-based feature maps. The feature maps include utilized routing resources by IP block, congestion metrics that include local/global net features and RUDY scores, and measures of the density of logic gate pins, clock pins, logic cells, and filler cells. Generative adversarial networks (GANs) are implemented in [15] to predict routing congestion. Using images of placed circuits placement and connectivity as input, the GAN generates routing heat maps that visually represent levels of congestion. The generator, also based on a U-Net model, creates heat maps while the discriminator ensures the generated heat maps are realistic by comparing the generated maps to actual data, refining the accuracy of the model through adversarial learning. Another approach enhances U-Net skip connections with a spatial attention module that integrates average-pooling, max-pooling, and standard deviation operations to address data imbalance [16]. In [17], a hyper-image based CNN is used to predict DRVs. The hyper-image is a multi-dimensional representation combining placement and global routing features, such as pin counts, neighborhood pin density, and routing overflow (horizontal and vertical), across multiple metal layers. The approach preserves spatial and layer-specific congestion characteristics, offering a more comprehensive input to the model than traditional 2-D congestion maps. DRVNet[18] DRV hotspot detection by utilizing early global routing (eGR)-derived features within a U-Net architecture to predict layer-wise DRV occurrences. Unlike traditional congestion models that rely on a single design snapshot, DRVNet extracts layer-wise congestion features from multiple eGR conditions. GCell-based congestion maps are incorporated to capture routing resource utilization, pin density, and via distribution across metal layers. Using the grid-structured predictions from DRVNet, BlkgComp[18], a classification-based blockage placement framework, employs ResNet-50 to identify congestion-prone regions and generate initial blockage sets for routing blockage placement. An RL agent further optimizes blockage placement using a policy function to adjust blockages and a reward function to balance DRV reduction and routing feasibility.

Images are typically used for routability prediction, but there are also techniques that utilize graphs. In [19], heterogeneous graphs are used to encode layout objects and any corresponding relationships. The graph represents global routing cells and nets as nodes, with edges capturing connections, overlaps, and neighboring relationships within the circuit. The technique uses a GCN to process the graph and predict routing demand and congestion status (binary classification) for each GCell. Beyond using only an image or graph in isolation, a hybrid model that combines a pin graph neural network (PGNN) and a U-Net is proposed in [20] to predict DRC hotspots. The approach integrates a pin proximity graph, capturing detailed spatial relationships and pin accessibility, with grid-based layout features including pin density, RUDY, routing capacity, and macro blockage data to represent a context of the global circuit design. The combined representation enables the model to accurately identify areas of the circuit considered DRC hotspots, addressing challenges in advanced technology nodes where pin accessibility and routing congestion heavily influence manufacturability.

### 3.2 Case II: Timing Prediction

ML is increasingly used in timing analysis to improve accuracy and efficiency across various IC design stages. ML models predict timing metrics early in the design process, accounting for parasitics, process variations, and aging effects. Early methods [34–38] utilized traditional ML algorithms, including support vector machines (SVMs) and random forests (RFs) or fully connected neural networks, to predict timing metrics such as delays and slew rates. The early algorithms often focused on specific stages of the design

**Table 1: Prior work utilizing representation learning in digital design automation.**

| Category | Ref | Objective | Problem Type | Modality | Representation/Feature Description | Network Architecture |
|---|---|---|---|---|---|---|
| Routability Prediction | [13] | Predict number of DRV and DRC hotspots during placement | Regression & Classification | Image | Pin density, macro regions, RUDY scores | ResNet for DRV and U-net for DRC hotspots |
| | [14] | Predict if a tile in the layout is a Design Rule Check (DRC) hotspot | Classification | Image | Pin configuration images and tile features (routing resources, congestion metrics, and density metrics) | Customized U-Net (J-Net) |
| | [15] | Predict routing congestion based on heatmaps generated from placement | Generative | Image | Post-placement layout images combined with connectivity images (maps netlist connections as edges based on placement locations) | GAN |
| | [16] | Predict routing congestion heatmaps during placement | Regression & Classification | Image | Grids of numerical features per tile: Macro Regions, RUDY, Pin RUDY, Cell Density, and Congestion Map. | U-Net enhanced with spatial attention |
| | [17] | Predict if a tile in the layout is DRV prone | Classification | Image | Hyper-images combining placement (pin counts, neighborhood pin density) and global routing (routing overflow by layer) | CNN |
| | [18] | Predict layer-wise DRV hotspots to optimize routing blockages | Classification & Optimization | Image | GCell-based congestion maps, multiple early global routing (eGR) runs, cell/macro density, pin density, routing track utilization, via counts | U-Net for DRV prediction, ResNet-50 for blockage ranking, Actor-Critic for blockage optimization |
| | [19] | Predict whether the routing demand exceeds the routing capacity of the GCell boundary | Classification | Graph | Heterogeneous Graph consisting of<br>- global cells/nodes with routing capacity, net/pin overlap density, and macro overlap as features<br>- net nodes with bounding box dimensions, net overlap density, and pin overlap count as features | Heterogenous GNN |
| | [20] | Predict DRC hotspots considering pin accessibility and routing congestion | Classification | Graph & Image | Pin proximity graph consisting of<br>- pins as nodes with pin location, orientation, cell type of pins (standard cell, macro, etc.), and proximity features (distance to nearest pins, etc.)<br>- proximity-based edges connecting pins that are within a defined distance<br>- edges linking pins with shared access points or overlapping regions<br>Grid-based features (e.g., Pin Density, RUDY, Routing Capacity, etc.) | Combined GNN (Pin Proximity Graph) and U-Net (Grid Features) |
| Timing Prediction | [21] | Predict post routing gate arrival time at post floorplan, placement, and CTS design stages | Regression | Graph | Timing path graphs with gates, inputs, and outputs as nodes with attributes including structural features (logic level, fan-out count, standard cell encoding, etc.), timing based features (stage delay, arrival time, etc.) and parasitics. | GNN |
| | [22] | Predict pre-routing arrival time and slack at timing endpoints | Regression | Graph | Heterogeneous graph with<br>- pins as nodes (features: pin capacitance, position, type)<br>- net/cellas edges (features: distances, delay, slew) | GNN |
| | [23] | Predict post routing wire slew and delay | Regression | Graph | RC network graph consisting of<br>- interconnect capacitance as nodes<br>- interconnect resistance as edges | GraphSAGE and tranformers |
| | [24] | Predict pre-routing endpoint arrival time considering routing-induced parasitics and timing optimizations | Regression | Graph | Heterogeneous graph<br>- cells as nodes with standard cell encoding, drive strength, slew, and slack as features<br>- pins as nodes with pin type, capacitance, and arrival time as feature<br>- cell arc as edges with cell delay and slew as features<br>- net arc as edges with net length and delay as features | GAT |
| | [25] | Predict instance-level timing hotspots and slack distributions at placement | Regression | Image | Mesh grid-based maps of transition time, instance density, and timing slack (worst slack, total slack, negative slack) | ResNet and U-Net |
| | [26] [27] | Predict net resistance, capacitance, arc delay, and slew at the pre-route stage for timing optimization | Regression | Graph & Image | Timing graph with<br>- pins or cells as nodes with electrical properties (e.g., capacitance, resistance), timing slack, arc delay, and slew as features<br>- timing arcs between nodes as edges<br>Image based congestion maps (e.g., RUDY, density, pin capacitance location) | GNN and CNN (multimodal) |
| Placement optimization | [28] | Global placement (analytical)<br>- Predict routing congestion of each standard cell, as a cost term of the objective function for a gradient-based optimization | Regression | Graph | A heterogeneous graph consisting of three subgraphs, with three node types,<br>- cell nodes with cell size, connectivity to nets, and cell geometry as features<br>- net nodes with span of nets and connectivity to cells as features<br>- layout grid-cell nodes with RUDY and center location of grid cells as features<br>(1) Topological subgraph (connectivity between cell-pin-net): a heterogeneous subgraph of cell and net nodes<br>  - with edge attributes representing pin and signal direction<br>(2) Geometrical subgraph (location of grid-cells, connectivity between grid cells): a homogeneous subgraph of grid-cell nodes<br>  - with edge attribute representing connectivity between grid cells<br>(3) Geometrical subgraph (location of cells with respect to grid cells): a heterogeneous subgraph of cell and grid-cell nodes<br>  - with edge attributes representing distances between cell locations and the center of a grid cell | GNN |
| | [29] | Macro placement<br>- Sequentially predict optimal placement location for each macro<br>- approximate policy function and value function of the RL agent | Optimization | Graph & Image | Macro topology: Homogeneous graph of macro nodes with<br>  - geometric information of the macro, e.g. size, number of pins, and types of pins as node feature<br>  - adjacency between two macros as edge<br>Image of the current partially placed canvas (binary matrix marking if the grid cell is occupied or not)<br>Netlist metadata vector, identity number (an integer) of the current macro being placed | Actor-critic architecture using CNN and GAT |
| | [30] | Macro placement<br>- Sequentially predict optimal placement location for each macro<br>- approximate policy function and value function of the RL agent | Optimization | Image | (1) Image of the current partially placed canvas (0/1/0.5 value marking if a grid cell is occupied, where 0.5 marks the edge boundary of the placed macros)<br>(2) Position mask matrix (binary value marking if a grid cell is a valid placement position for the bottom-left corner of the macro currently being placed)<br>(3) Wire mask matrix (value of increased HPWL if the current macro is placed at the grid cell position) | Actor-critic architecture using customized U-net |
| | [31] | Macro placement<br>- Sequentially predict an optimal placement location for each macro<br>- approximate an option-selecting policy function; and under each option, an action-selecting policy function and a termination probability function, of the HRL agent | Optimization | Graph & Image | Image of the current partially placed canvas (binary matrix marking if the grid cell is occupied or not)<br>Macro netlist graph, with macro as node<br>  - currently-being-placed status flag as node feature<br>  - connectivity between macros as edge | Option-critic architecture using CNN and GNN |
| | [32] | Optimize placement to improve power, performance, and area (PPA); reduce congestion, timing delays, and power consumption | Clustering | Graph | Netlist graph with<br>  - cells as nodes with features such as position, timing, power attributes | GraphSAGE and K-means clustering |

flow. In the last decade, neural network-based graph methods have gained prominence in timing analysis due to the natural representation of circuit topologies as graphs.

GNNs excel at capturing dependencies and topological relationships, using message-passing methods to model cumulative delays and interdependencies, providing a powerful tool for timing-based ML in IC design. In [21], timing paths defined in static timing analysis (STA) reports are represented as graphs and applied to a GCN to predict the arrival time of a signal to the gates of a timing path. The post-routed arrival time is predicted in early design stages (floorplan, placement, CTS). The nodes in the timing path graphs contain features that include the delay of the gate, the arrival time provided at the current design stage, the fan-out, the logic level, the estimated interconnect length, and the interconnect capacitance. A GNN-based method is introduced in [22] to predict the pre-routing arrival time and slack at timing endpoints. The circuit is modeled as a heterogeneous graph, where pins are represented as nodes, with features that include capacitance, distance to die boundaries, and timing endpoint indicators, and nets are represented as edges, with attributes that include length, delay, and slew from LUTs. The GNN utilizes a level-by-level message-passing technique to emulate the propagation of timing paths in STA. GNNTrans, a GraphSAGE-based network adapted for wire delay and slew estimation, is introduced in [23] to model interconnect *RC* networks as graphs. Unlike previous methods that target timing paths or heterogeneous circuit elements, GNNTrans embeds *RC* graphs by aggregating local node features (e.g., resistance, capacitance) and global path features (e.g., Elmore delay, D2M delay). Attention mechanisms assign weights to neighboring nodes and edges based on relative importance, allowing the model to prioritize critical nets and timing paths. In [24], the circuit netlist is modelled as a heterogeneous graph with nodes (cells, pins) and edges (cell arc, net arc, gate) that captures intra-cell and inter-cell timing relationships. A heterogeneous graph attention network (HGAT) [39] processes the graph, using attention mechanisms to prioritize critical interactions (e.g., pin-to-cell effects) and residual connections to prevent over-smoothing. The HGAT enables effective learning of local patterns and global dependencies, which ensures an accurate pre-routed prediction of timing.

While graph-based methods excel at leveraging the topological nature of circuits for timing prediction, image-based methodologies offer a complementary set of techniques by treating timing analysis as a spatial problem. In [25], a Res-UNet is used to process image-like maps, including transition time and timing maps (e.g., WNS and TNS), to predict post-routing timing distributions and identify critical timing hotspots. By utilizing spatial circuit characteristics, the image-based technique offers an alternative method to analyze and optimize circuit timing. The potential to combine graph and image modalities to provide a multimodal approach for timing analysis is highlighted in [26] and [27]. GNNs and CNNs are combined in [26] to improve pre-route timing prediction. GNNs model circuit dependencies to predict net resistance and capacitance, while CNNs process spatially-organized features such as congestion maps to refine arc length predictions. The multi-modal network proposed in [26] is integrated into a commercial EDA tool in [27], where the predicted timing predictions are used to replace default estimates, thereby improving WNS, TNS, and reducing timing violations.

## 3.3 Case III: Automated Placement

Macro placement and global placement are similar optimization tasks that aim to optimally place circuit macros and standard cells, respectively, while aiming to reduce total wirelength, routing congestion, and cell density. However, the scales of the problem differ as there are many fewer macros than there are standard cells in a typical circuit. Consequently, deep reinforcement learning (RL) has been applied to macro placement [29–31], while for global placement, machine learning has been applied to assist existing gradient-based analytical placement algorithms [28]. However, both techniques utilize deep neural networks to approximate mathematical functions within traditional optimization algorithms, which are often hard to derive. For deep RL, neural networks are used to approximate/learn a policy function and a value function of an actor-critic RL agent [29, 30]. In addition, an option-selecting policy function, an option-specific action-selecting policy function, and a termination function for each option of an option-based hierarchical RL (HRL) agent are approximated in [31]. In [28], a neural network is used to approximate a cost term in the objective function of an analytical placement algorithm. Clustering approaches that utilize PPA metrics for placement optimization have also been explored [32]. In this section, the input features (geometrical and topological) and the network architectures of ML-based placement algorithms are analyzed to compare across different representation learning approaches.

Geometrical input features are required to represent the circuit layout of the components placed (macros and standard cells). When deep RL is used for macro placement, the layout with the partially placed macros represents the current state of the RL optimization algorithm. SRLPlacer [29], HRLP [31] and EfficientPlace [30] use an $N \times N$ matrix to represent the placement positions on grid sizes of $20 \times 20$, $32 \times 32$, and up to $512 \times 512$, respectively, where a binary value is used to flag whether a grid cell is occupied. A binary matrix, which can be represented as an image, is used as a geometrical input to a neural network, where a CNN is used in [29] and [31] and a U-net is used in [30]. EfficientPlace also uses two mask matrices as image inputs to a U-net, as listed in Table 1. For global placement of standard cells, RoutePlacer [28] uses two subgraphs to represent layout information. A homogeneous subgraph of interconnected grid-cells is used to map the tiles of a layout onto a graph. A second heterogeneous subgraph of grid-cell and standard-cell nodes is used to retain the placement location of the standard cells with respect to the grid-cells.

Netlist and topological information is naturally represented using graphs, for both macro and standard cell placement. A graph of a netlist of macros is used in SRLPlacer [29] and HRLP [31]. However, EfficientPlace [30] uses macro netlist connectivity and macro area as features to sort the placing order of the macros, and does not use the netlist features directly as inputs to the U-net. In RoutePlacer [28], the netlist subgraph contains connectivity information between cells, pins, and nets, which further emphasizes on net features beyond just accounting for standard cell features. The netlist subgraph used in RoutePlacer provides much richer information on the circuit topology as compared to the macro netlist graphs in SRLPlacer and HRLP.

For embedding, various approaches are analyzed. For deep RL, EfficientPlace [30] takes only images as inputs and utilizes a U-net for the embedding. In contrast, SRLPlacer [29] and HRLP [31] are able to take both graphs and images as input by concatenating the netlist embedding and layout embedding. SRLPlacer uses GAT, while HRLP uses GNN to process graph inputs into netlist embeddings, and both utilize a CNN to process image inputs to generate layout embeddings. RoutePlacer [28] uses a GNN with message passing to process one primary graph consisting of three subgraphs, and utilized max pooling and concatenation to fuse messages together.

A PPA-directed clustering framework is introduced in [32] that optimizes placement by modeling netlists as graphs, where cells are represented as nodes and connections as edges. Each node is populated with physical, timing, and power attributes. The framework integrates similarity-driven optimization, which groups cells within the same net while minimizing distant connections, and PPA-directed optimization to improve metrics on power, performance, and area. Clustering is performed on GraphSAGE-generated embeddings, using K-means and guided by loss functions to align clusters, reduce congestion, and enhance timing and power efficiency.

## 4 FUTURE DIRECTIONS

This section outlines future directions for advancing representation learning in EDA. The areas of interest include multimodal learning, natural language modality and large language models (LLMs), and workflow standardization, which are discussed in sections Section 4.1, Section 4.2, and Section 4.3, respectively.

### 4.1 Multimodal Learning and Alignment

Multimodal learning [40, 41] is gaining traction in EDA as the use of diverse data types including graphs and images provides a more holistic representation of circuit features. Recent work, including [19], [26], and [27], highlight the advantages provided when combining graph and image modalities. The proposed techniques utilize the complementary strengths of both graph and image representations to provide improved prediction accuracy and optimization efficiency. Future research on systematic ablation studies, as suggested in [42], is imperative to quantify the individual contributions of each modality to the overall performance of the model.

Another compelling use case for multimodal learning in EDA is modality alignment, where one modality is used to guide or enhance the understanding of another [43, 44]. For example, alignment techniques allow for image representations to provide spatial or topological context for graph-based analysis, or vice versa. Alignment of multimodal representations improves tasks including feature extraction, anomaly detection, or design optimization by bridging gaps in information and resolving ambiguities inherent to individual modalities. Several intersections in features between graph and image/grid modalities are highlighted in Table 1. As an example, the RUDY metric is treated as a grid-based feature in [16], while being utilized as a graph-based feature in [20]. The overlapping use of features illustrates an opportunity for modality alignment, where the shared aspects of RUDY are utilized and linked across modalities to provide a unified representation. By aligning such features, multimodal systems reduce redundancy and enhance the depth of analysis.

### 4.2 Natural Language Modality and Large Language Models

In EDA, text-based data is predominantly derived from register transfer level (RTL) descriptions, which serve as a foundational representation of the circuit. Beyond RTL, other valuable text-based data resources, including textbooks, research papers, and technical documentation, also provide significant insights. Despite the availability of such data, text modality remains underutilized in EDA, representing an area of untapped potential to improve automation and optimization.

The emergence of LLMs has introduced new opportunities to utilize text-based data in circuit design applications. The LLMs excel at interpreting, generating, and analyzing text, which enables tasks such as translating natural language requirements into formal circuit design specifications, extracting key insights from text-based resources, and automating the process of documentation [45]. Current applications of LLMs in EDA include generating HDL code from high-level descriptions[46–48], interpreting code and analyzing logs for debugging and verification [49–51], and facilitating chatbots that assist with design queries, provide insight of product documentation, and guiding through tool usage [52, 53]. However, the integration of LLMs with other data modalities, such as graphs and images, remains relatively unexplored. Combining the language processing and semantic understanding capabilities of LLMs with spatial and relational insights from graph and image-based data has the potential to create more comprehensive and adaptive EDA flows.

### 4.3 Standardizing EDA through Open Data and Frameworks

Multimodal learning and alignment in EDA depend on unified data formats, workflows, and evaluation protocols to ensure comparability, reproducibility, and reusability. While EDA research frequently utilizes overlapping features and modalities—such as netlist graphs for structural analysis, layout images for spatial insights, and timing data for performance evaluation—disparities in datasets, PDKs, and preprocessing methods create inefficiencies and hinder benchmarking and model reuse. In addition, publication or open profiling of datasets generated using commercial EDA tools and PDKs is often restricted by terms of use requirements, further complicating efforts to share and standardize data. Unified frameworks and standardized data representations address these challenges by enabling seamless integration of diverse modalities, fostering alignment, and allowing hybrid approaches that leverage complementary strengths. Such efforts promote collaboration, accelerate innovation, and enable scalable and efficient EDA pipelines, transforming how multimodal data is utilized in representation learning.

To address such challenges, several initiatives have focused on creating a unified infrastructure for EDA and ML-driven workflows. Efforts include the development of standardized data representations and open datasets to support consistent benchmarking [54–56], advancements in ML infrastructure [57, 58], and the proliferation of open-source design toolsets [59, 60] and resources

Deep Representation Learning for Electronic Design Automation

[61, 62]. Such initiatives provide a strong foundation for future work in standardizing EDA workflows and fostering integration of multimodal learning techniques.

## 5 CONCLUSION

This paper explores the current landscape and potential of deep representation learning in EDA. The paper highlights the ability of deep representation learning to leverage diverse modalities including graphs, images, and hybrid approaches to address challenges in modern circuit design. By categorizing key applications in Table 1 and analyzing general use cases in routability prediction, timing analysis, and automated placement, the utilization of deep representation learning to improve prediction accuracy, optimization efficiency, and scalability is demonstrated. Future directions on expanding multimodal learning, enhancing interoperability, and exploring new methods for aligning diverse data modalities are discussed, which enable scalable and robust solutions to meet the growing complexity of circuit design workflows.